\documentclass[letterpaper, 10 pt, conference]{ieeeconf}  

\IEEEoverridecommandlockouts                              
\usepackage[utf8]{inputenc}
\usepackage[T1]{fontenc}
\usepackage{graphicx}
\usepackage[namelimits]{amsmath} 
\usepackage{amssymb}             
\usepackage{amsfonts}            
\usepackage{mathrsfs}            
\usepackage{subfigure} 
\usepackage{booktabs}
\usepackage{cite}
\usepackage{url}
\usepackage{dsfont}
\usepackage{algorithm}
\usepackage{algpseudocode}
\usepackage{algorithmicx}
\usepackage{amsmath}
\usepackage[switch]{lineno}
\usepackage{bbding}

\usepackage{hyperref}

\usepackage{multirow}
\usepackage{textcomp,booktabs}
\usepackage[usenames,dvipsnames]{color}
\usepackage{colortbl}
\definecolor{mgray}{gray}{.9}
\hypersetup{
    colorlinks=true,
    linkcolor=blue,
    filecolor=magenta,      
    urlcolor=cyan
}

\usepackage[colorinlistoftodos,bordercolor=blue,backgroundcolor=blue!20,linecolor=blue,textsize=scriptsize]{todonotes}
\title{\LARGE \bf
STrajNet: Multi-modal Hierarchical Transformer for Occupancy Flow Field Prediction in Autonomous Driving}

\author{Haochen Liu, Zhiyu Huang, and Chen Lv$^{*}$,~\IEEEmembership{Senior Member, IEEE}
\thanks{Code will be available at: {\href{https://github.com/georgeliu233/STrajNet}{https://github.com/georgeliu233/STrajNet}}}
\thanks{H. Liu, Z. Huang, and C. Lv are with the School of Mechanical and Aerospace Engineering, Nanyang Technological University, 639798, Singapore. (E-mails: {\tt \{haochen002, zhiyu001\}@e.ntu.edu.sg, lyuchen@ntu.edu.sg})}
\thanks{This work was supported in part by the A*STAR Project (W1925d0046) and the SUG-NAP Grant (No. M4082268.050) of Nanyang Technological University, Singapore.}
\thanks{$^{*}$Corresponding author: C. Lv}
}

\begin{document}
\maketitle
\thispagestyle{empty}
\pagestyle{empty}

\begin{abstract}
Forecasting the future states of surrounding traffic participants is a crucial capability for autonomous vehicles. The recently proposed occupancy flow field prediction introduces a scalable and effective representation to jointly predict surrounding agents' future motions in a scene. However, the challenging part is to model the underlying social interactions among traffic agents and the relations between occupancy and flow. Therefore, this paper proposes a novel Multi-modal Hierarchical Transformer network that fuses the vectorized (agent motion) and visual (scene flow, map, and occupancy) modalities and jointly predicts the flow and occupancy of the scene. Specifically, visual and vector features from sensory data are encoded through a multi-stage Transformer module and then a late-fusion Transformer module with temporal pixel-wise attention. Importantly, a flow-guided multi-head self-attention (FG-MSA) module is designed to better aggregate the information on occupancy and flow and model the mathematical relations between them. The proposed method is comprehensively validated on the Waymo Open Motion Dataset and compared against several state-of-the-art models. The results reveal that our model with much more compact architecture and data inputs than other methods can achieve comparable performance. We also demonstrate the effectiveness of incorporating vectorized agent motion features and the proposed FG-MSA module. Compared to the ablated model without the FG-MSA module, which won \emph{$2^{nd}$ place} in the 2022 Waymo Occupancy and Flow Prediction Challenge, the current model shows better separability for flow and occupancy and further performance improvements.
\end{abstract}


\section{Introduction}
\label{sec1}
Making robust and accurate predictions of future motions for multiple traffic participants (agents) in an efficient and scalable manner is one of the core capabilities of autonomous vehicles (AVs) \cite{mozaffari2020deep, huang2021driving, huang2022differentiable}. However, motion prediction is an extremely difficult task due to a handful of challenges. First, the AV has to confront countless complicated traffic scenes, which consist of a varying number of heterogeneous traffic elements \cite{mo2022multi}. Second, the motion predictor should deal with not only the existing traffic elements' observed states but also the underlying complicated interactions among them \cite{mo2020interaction}. In addition, the prediction results are required to be robust to handle uncertainties \cite{djuric2020uncertainty}, as the agent's future motions may differ vastly even under the same situations.

\begin{figure}[htp]
    \centering
    \includegraphics[width=\linewidth]{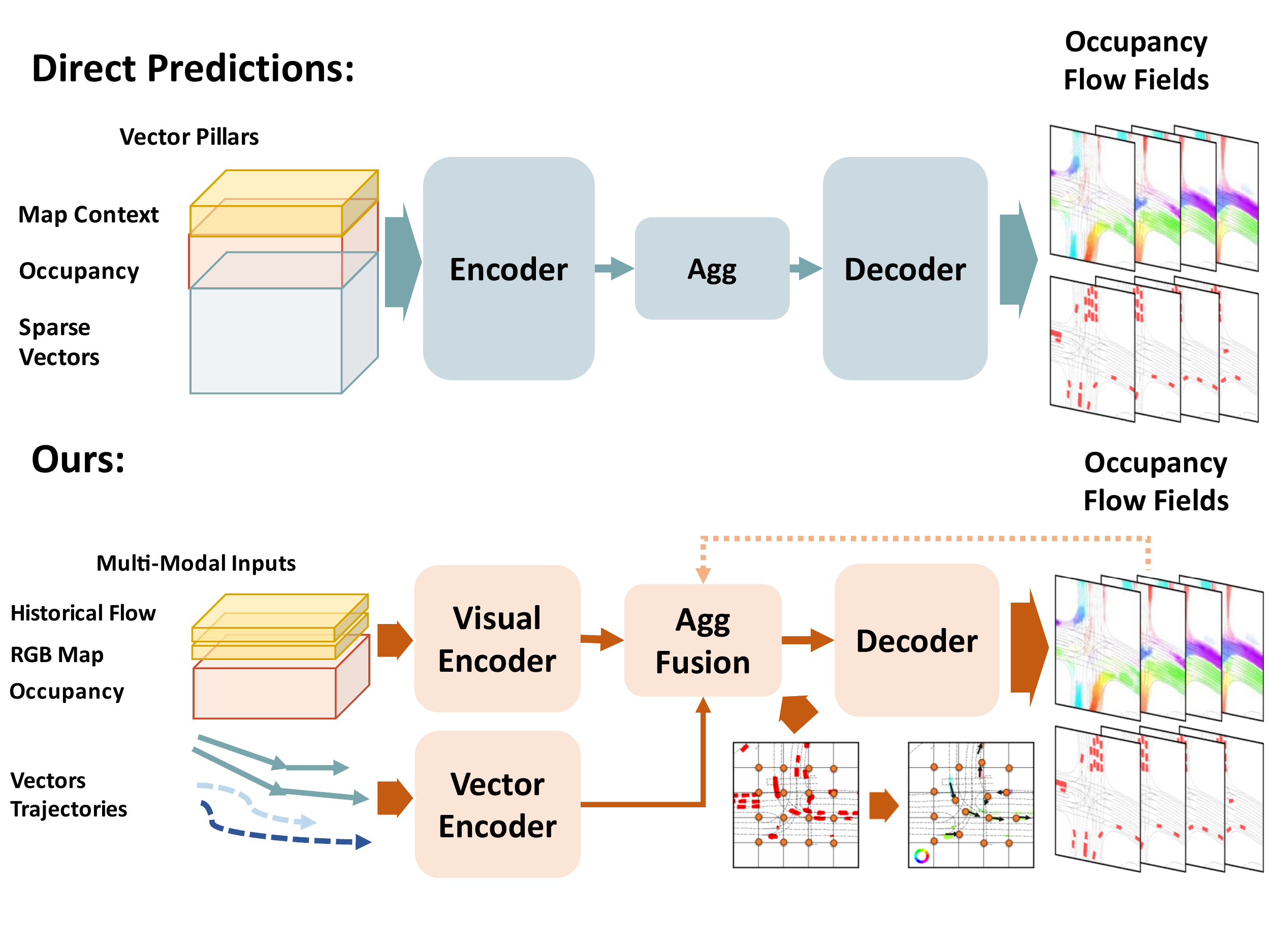}
    \caption{Illustration of our method and direct occupancy flow field prediction method. Compared to direct prediction frameworks \cite{mahjourian2022occupancy, hu2022hope, hoermann2018dynamic}, our method leverages a late fusion of visual and vector features, as well as the proposed flow-guided attention. Our method has much more succinct data inputs and model parameters but outperforms similar methods.}    
    \label{fig0}
\end{figure}

While most existing works on multi-agent prediction directly predict a sequence of future locations for each agent \cite{ngiam2021scene, varadarajan2022multipath++, gu2021densetnt}, the recently proposed occupancy flow field prediction \cite{mahjourian2022occupancy} offers a more efficient and scalable prediction representation for multiple agents in a scene \cite{kim2022stopnet}. Specifically, the occupancy flow field is a spatial-temporal grid \cite{hoermann2018dynamic} that 1) simultaneously forecasts the occupancy probabilities for all agents (observed and occluded) on the grid, and 2) outputs a set of backward occupancy flow that predicts the warping (pixel shifting) between each occupancy grid for each occupied grid cell. It provides a competitive alternative prediction representation with better scalability and efficiency due to its ability to predict a varying amount of agents at once, better safety for downstream decision-making thanks to occluded (speculative) agent predictions, as well as dynamic tractability for each agent by the backward flow. However, current occupancy flow field prediction frameworks \cite{mahjourian2022occupancy, hu2022hope} based on pillar-inspired \cite{lang2019pointpillars} inputs are redundant and consume a large number of memory resources. Moreover, the direct prediction method (see Fig. \ref{fig0}) lacks separability for flow and occupancy features, which are mathematically related.  

To tackle this challenge, we propose a multi-modal hierarchical Transformer-based framework for the spatial-temporal prediction task. First, a Swin-Transformer \cite{liu2021swin} visual encoder is utilized to fuse the information of visual modalities and capture the interactions among historical occupancy, backward flow, and dense road map. We also employ the vectorized representation of agent historical motions and encode them through a Transformer-based vector encoder considering interaction awareness. To model the mathematical relationships between future occupancy grids and backward flow, we design a flow-guided multi-head self-attention (FG-MSA) module for the high-level encoded grid cell that queries the warped grid cell by learnable flow offsets. It then outputs concatenated flow and occupancy grid cells for all future steps. To better associate occupancy grid cells with corresponding agent trajectories to capture their motion tendencies, we employ a cross-attention module that queries the encoded interaction trajectories of each grid cell across future steps. It should be noted that our proposed hierarchical Transformer framework is concise with simply three encoding stages, yet achieves state-of-the-art performance. The contributions of our proposed framework are summarized as:
\begin{enumerate}
\item We propose a novel multi-modal hierarchical Transformer framework for the occupancy flow prediction task. It can fuse the rasterized and vectorized features of the scene and capture the underlying interactions.
\item We design a flow-guided attention module that effectively queries the grid cell features warped by learnable flow offset, which shows better separability.
\item We validate the framework on a large-scale real-world driving dataset, and the proposed model with a concise structure achieves state-of-the-art performance.
\end{enumerate}

\section{Related Work}
\subsection{Transformers for motion prediction}
Due to the computational efficiency and effectiveness of the multi-head attention mechanism in time-series or graph-like interaction encoding, Transformer-based structures have gained great success in motion prediction tasks. Transformers have been widely adopted in vectorized scene encoding of both historical agent trajectories and high-fidelity map segments \cite{gao2020vectornet, huang2022multi}. On the other hand, with the help of the vision Transformer (ViT) \cite{dosovitskiy2020image}, rasterized scene features can also be efficiently encoded with larger visual reception fields compared to CNN-based methods. In our method, we further utilize those two categories of scene representations during encoding and combine them using different types and stages of Transformer encoders, which shows improved performance. Another important factor for motion prediction is interaction modeling among the traffic agents and map information, and Transformers can well handle the interaction graphs through the attention mechanism. For example, VectorNet \cite{gao2020vectornet} firstly proposed to model the traffic elements as a fully-connected graph. SceneTransformer \cite{ngiam2021scene} further unified the interaction modeling across and within vectorized map and agent history sequences. In our work, we consider the interaction among occupancy grid cells and among the vectorized agent motions across different time steps, using self-attention Transformers modules to process them individually, as well as the interaction between these two kinds of modalities with cross-attention Transformer modules. Moreover, inspired by DAT \cite{xia2022vision} in object detection, we propose a flow-guided attention mechanism that is suitable for incorporating both flow and occupancy grid predictions, so that the grid cell features can be adaptively queried from the guidance of flow offsets.


\subsection{Occupancy flow field for motion prediction}
Forecasting the future motions through occupancy grids can date back to ChauffeurNet \cite{bansal2018chauffeurnet}, which predicts the future occupancy map to perform behavior planning in autonomous driving. StopNet \cite{kim2022stopnet} further utilizes this objective and improves the overall scene representation for efficient motion prediction. However, they ignore the tractability between time steps in the future, and \cite{mahjourian2022occupancy} addresses this issue and makes the dynamic feature of each grid cell tractable by predicting the backward flow. Following the same representation, HOPE \cite{hu2022hope} proposed a hierarchical spatial-temporal predictor comprising a deep-level multi-stage encoder-decoder, as well as 3 layers of aggregators \cite{hu2021fiery} for fusing high-level visual features. However, using the traditional spatial-temporal hierarchical structure costs a deeper model and much more model parameters, and requires pretraining the visual encoder \cite{liu2021swin}. Concurrent to our work, VectorFlow \cite{huang2022vectorflow} instead fuses vector and visual features using cross-attention with a simple CNN-based encoder-decoder. However, it involves too much redundancy in vectorized inputs and the visual features are encoded with small reception fields. Different from the methods above, we instead design a more compact hierarchical Transformers structure that can deal with multi-modalities fusion with only 3 stages of encoding. In addition, a single layer aggregation of proposed flow-guided self-attention can also help improve the multi-task learning process and boost the final performance.

\section{Methodology}
\subsection{Problem Formulation}
Forecasting the occupancy flow field can be formulated as concurrently predicting a multi-task output $\hat{\textbf{Y}}$ that consists of the critical future frames of observed occupancy $\hat{O}_{k}^{b}$, occluded occupancy $\hat{O}_{k}^{c}$, and corresponding backward flow $\hat{F}_{k}$ at each future step $k \in [1, T_f]$, conditioned on the past and current states of traffic agents and scene context inside a certain area. More specifically, the occupancy grid is modeled as a binary single-channel image $\hat{O}_{k}^{b},\hat{O}_{k}^{c}\in R^{H\times W\times 1}$, where $\hat{O}_{k}^{b}$ represents the future occupancy of currently observed agents, and $\hat{O}_{k}^{c}$ denotes the occluded ones that might occur in the future. The backward flow is regarded as a two-channel image-like tensor for the motion shifting of grids by occupied traffic agents along $x$ and $y$ axis: $\hat{F}_{k}=(x, y)_{k-1}-(x, y)_{k}  \in R^{H \times W \times 2}$. The values of $\hat{F}_{k}$ range from $\pm H/2,\pm W/2$ along each axis. Mathematically, given the warping function $f_{\mathcal{W}}$, flow-warped occupancy is defined as:
\begin{equation}
\label{e1}
O_k = f_{\mathcal{W}}(O_{k-1},F_{k-1})\odot O_k.
\end{equation}

For the input representation $\textbf{X}$, it comprises multiple modalities with detailed formulations as follows:

\begin{figure*}[ht]
    \centering
    \includegraphics[width=0.94\linewidth]{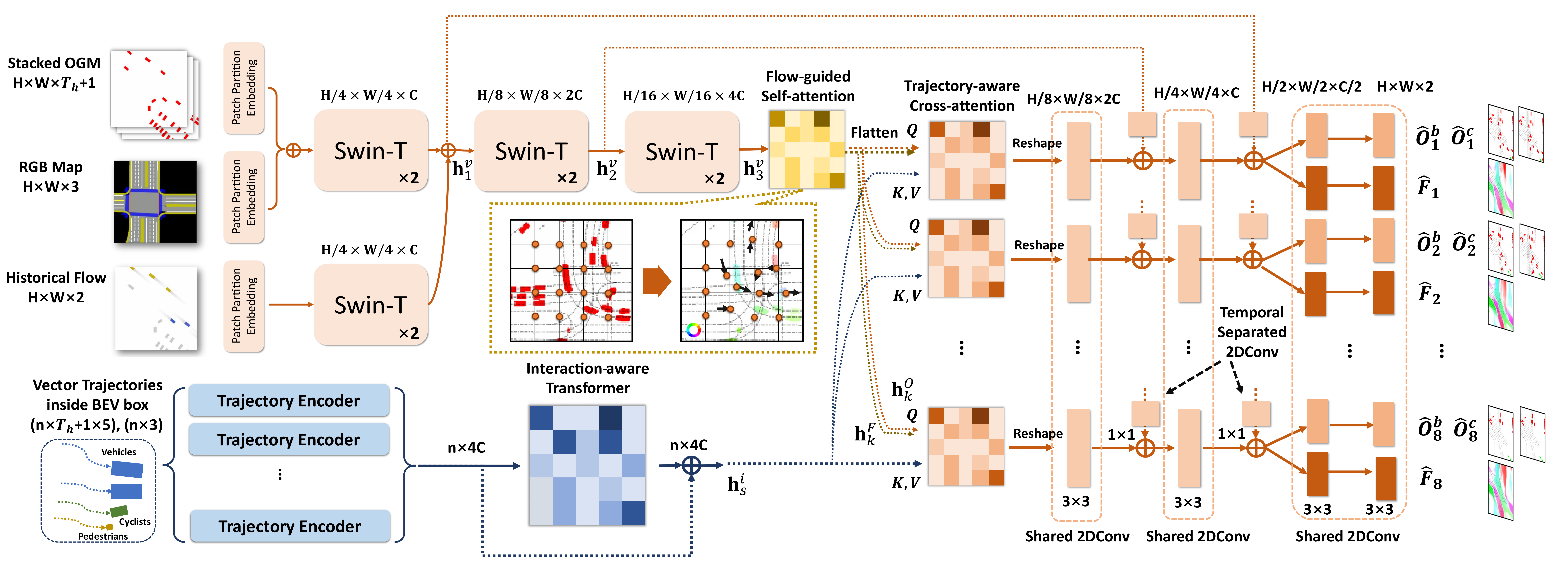}
    \caption{An overview of STrajNet. Multi-modal inputs are separately encoded by Swin-Transformer encoders and trajectory encoders; trajectory-based interaction awareness of each grid cell is encoded by cross-attention; flow and occupancy features are separated through flow-guided self-attention and decoded by a shared feature pyramid decoder to predict occupancy and flow for each timestep.}
    \vspace{-0.2cm}
    \label{fig:fig.1}
\end{figure*}

\begin{figure}[ht]
    \centering
    \includegraphics[width=\linewidth]{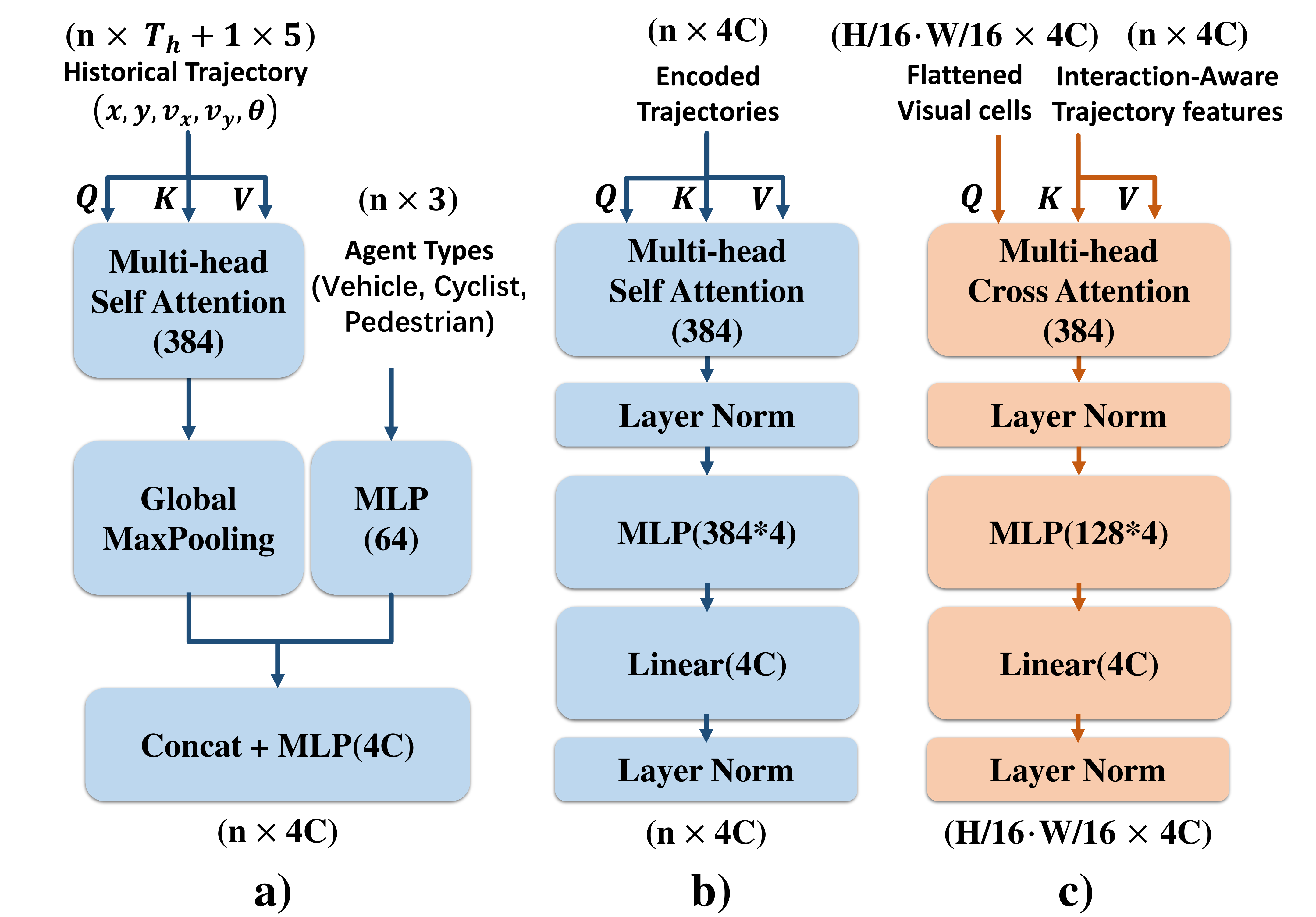}
    \caption{Detailed structures of a) Trajectory Encoder; b) Interaction-aware Transformer; c) Trajectory-aware Cross-attention.}
    \label{fig3}
\end{figure}

\textbf{1) Visual features:} To acquire the spatial-temporal occupied status for traffic agents, we firstly build the historical and current occupancy grid $O_t, t \in[-T_h, 0]$; dense road map $\mathcal{M}$ that renders road networks (colored by categories) and traffic light states as a rasterized RGB image \cite{huang2022recoat}. To facilitate the flow prediction, we also provide the historical backward flow $F_h$ built upon agent displacements in the occupancy grids between time steps: $-T_h,0$.

\textbf{2) Vector features:} For vectorized inputs, $n$ traffic agents currently occurred inside the grid area are collected to form a set of their historical trajectories $\mathcal{S}=\{S_1, S_2, \cdots, S_n\}$, each denotes a motion sequence $s_{t}^{i} \in S_i$ and each motion states is composed of $s_{t}^{i}=(x, y, v_x, v_y, \theta)$. We also concatenate the one-hot encoding of the traffic agent's type (vehicle, cyclist, or pedestrian) to the state. All of the representations above are normalized according to the current state of the ego vehicle. 

Suppose a prediction model $f$ with parameters $\theta$, the occupancy flow field prediction task is formulated as:
\begin{equation}
\label{e2}
\begin{aligned}
\hat{\mathbf{Y}} &= f(\mathbf{X}|\theta), \\
\mathbf{X} &= [\{O_t| t \in [-T_h, 0]\}; \mathcal{M}; \mathcal{S}; F_h], \\
\hat{\mathbf{Y}} &=\{(\hat{O}_{k}^{b}, \hat{O}_{k}^{c}, \hat{F}_{k})| k \in [1, T_f]\}.
\end{aligned}
\end{equation}

\subsection{Model Framework}
Fig. \ref{fig:fig.1} shows the overall structure of our model, which encompasses three fundamental modules for occupancy flow field prediction. First, the multi-modalities inputs are separately encoded. Vectorized trajectories $\mathcal{S}$ are encoded with trajectory encoder and interaction-aware Transformer to obtain the latent features $\{\textbf{h}_i^{s} | i \in [1,n]\}$. Visual features $O_t, \mathcal{M}, F_h$ are early-fused and encoded into latent features with multiple scales $[\textbf{h}_1^{v},\textbf{h}_2^{v},\textbf{h}_3^{v}]$ through a Swin Transformer-based \cite{liu2021swin} encoder. Then, we adopt the proposed flow-guided attention to aggregate the flow and occupancy features from the highest-level visual latent feature $\textbf{h}_3^{v}$. The obtained feature after the flow-guided attention layer is flattened and serves as queries to a temporally separated cross-attention module with $\textbf{h}_i^{s}$ as keys and values. Finally, a temporal-shared deconvolutional decoder with residual connections outputs a joint prediction of occupancy and flow $\hat{\mathbf{Y}}$.

\subsection{Multi-modal Encoder}
\textbf{1) Visual Encoder:} The multi-modal visual inputs are encoded by a Swin-Transformer-based encoder, and a separate Swin-Transformer block is set for historical flow $F_h$ for information ``shortcut" directly to future flow prediction. The occupancy map $O_t$, dense road map $\mathcal{M}$, and flow $F_h$ are initially embedded and down-sampled into shape $H/4 \times W/4 \times C$ by separated $4 \times 4$ convolution kernels with a stride of 4. We follow a concise setting of Swin-Transformer \cite{liu2021swin}.  More specifically, each Swin-Transformer module is a two-layer Transformer with both window self-attention (W-SA) and shifted window self-attention (SW-SA). It enables global and intersected attention-based interaction modeling for visual features. Each attention module is multi-head attention with relative positional encoding bias $\textbf{B}$: 
\begin{equation}
\label{e0}
\begin{aligned}
\operatorname{MSA}(\mathbf{Q}, \mathbf{K}, \mathbf{V}) = (\operatorname{head}_{1}|| \cdots||\text{head}_{\textbf{m}})W^{O},\\
\operatorname{head}_{i}=\operatorname{softmax}(\mathbf{Q K}^{T}/\sqrt{d}+\textbf{B}) \mathbf{V},
\end{aligned}
\end{equation}
where the head numbers are $\textbf{m}=[3,6,12]$ as the module goes deeper, and $d$ is the dimension of the key token. The visual encoder outputs a set of visual features in varying scales: $[\textbf{h}_v^{1},\textbf{h}_v^{2},\textbf{h}_v^{3}]$.

\textbf{2) Vector Encoder:} Vectorized trajectories are encoded considering interaction awareness. Historical motion vectors for $n$ agents are firstly aggregated across time by a shared trajectory encoder (Fig. \ref{fig3}(a)). It is a 4-head self-attention layer with global max-pooling, concatenated with its embedded agent type through a MLP layer. Next, a 6-head self-attention layer (Fig. \ref{fig3}(b)) with residual connection is introduced to build an interaction graph among all agents and outputs $\{ \textbf{h}_i^{s}|i\in [1,n]\}$. The latent dimensions are kept the same as $4C$ for all layers. 

\subsection{Aggregation and Fusion}
\label{FG}
\textbf{1) Flow-guided Multi-head Self-attention (FG-MSA):} To better aggregate the flow and occupancy features, we designed the FG-MSA module. The core idea is to use learnable flow offsets across future time steps to guide the flow-warped occupancy features via the MSA mechanism at once, so that both flow and flow-warped occupancy features for all future steps can be simultaneously aggregated. 

\begin{figure}[htp]
    \centering
    \includegraphics[width=\linewidth]{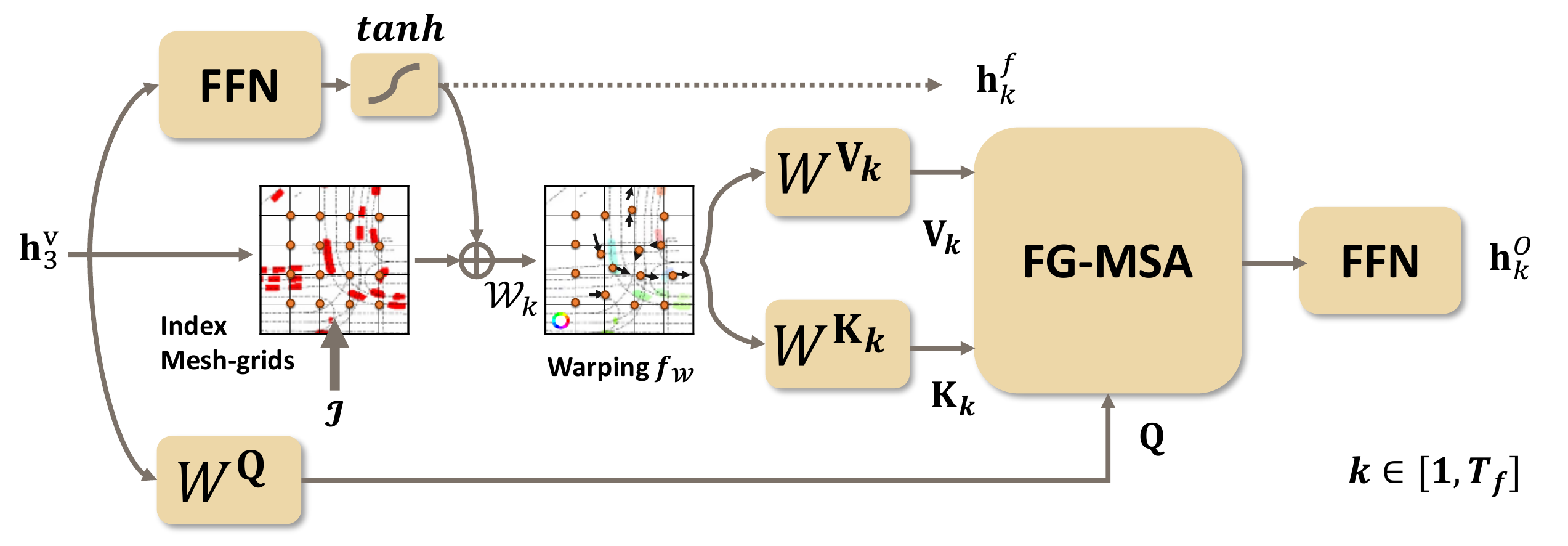}
    \caption{Computational pipeline of the proposed FG-MSA. Given $\textbf{h}_3^{v}$, generated flow offsets $\textbf{h}_k^{f}$ for each future time step are used to warp and guide the occupancy features $\textbf{h}_k^{O}$ through the MSA mechanism.}    
    \label{fig4}
\end{figure}

Given the latent visual feature $\textbf{h}_3^{v}$, flow offsets are projected using the tanh-normalized feed-forward network (FFN): $\textbf{h}_k^{f} = \tanh(\text{FFN}(\textbf{h}_3^{v})) \in R^{H/16\times W/16\times2}$. Then, given the index mesh-grids $\mathcal{I}$, the warped indices become $\mathcal{W}_k= \mathcal{I} + \textbf{h}_k^{f} = [\mathcal{W}_{k,x}; \mathcal{W}_{k,y}]$. We use bilinear interpolation as the warping function: 
\begin{equation}
\label{e4}
f_{\mathcal{W}}(\mathbf{X},\mathcal{W}_k)=\sum_{y,x}m(\mathcal{W}_{k,x},x)m(\mathcal{W}_{k,y},y)\textbf{X}_{y,x},
\end{equation}
where $m(i,j) = \operatorname{max}(0, 1-|i-j|)$. Then, the key and value are projected from flow-guided queries:
\begin{equation}
\label{e5}
\mathbf{K}_k,\mathbf{V}_k= f_{\mathcal{W}}(\textbf{h}_3^{v},\mathcal{W}_k)W^{\mathbf{K}_k,\mathbf{V}_k}, \ \mathbf{Q}= \textbf{h}_3^{v}W^{\mathbf{Q}}.
\end{equation}

Finally, we utilize an MSA module to enable each head to deal with each future timestep:
\begin{equation}
\label{e6}
\begin{aligned}
\operatorname{FG-MSA}(\mathbf{Q}, \mathbf{K}_k, \mathbf{V}_k) = \operatorname{head}_{k}W_k^{O},\\
\operatorname{head}_{k} = \operatorname{softmax}(\mathbf{Q K}_k^{T}/\sqrt{d}+\textbf{B}) \mathbf{V}_k.
\end{aligned}
\end{equation}

The flow-guided occupancy features $\textbf{h}_k^{O}$ are then obtained through a standard FFN.

\textbf{2) Trajectory-aware Cross-attention:} The purpose of this module is to associate each grid cell with trajectory information of presented agents, so that the information is more directed and no longer constrained by patches or nearby features. As shown in Fig. \ref{fig3}(c), we use the aggregated occupancy and flow features $\textbf{h}_k^{O} + W^{f}_k\textbf{h}_k^{f}$ as queries, and keys and values are vectorized agent motion features $\textbf{h}_i^{s}$. We implement 8 cross-attention modules for different future timesteps and the outputs are reshaped back to original shape of the visual feature.

\subsection{Decoder}
A feature pyramid network (FPN) \cite{lin2017feature} decoder is employed to decode occupancy and flow from the fusion feature. As shown in Fig. \ref{fig:fig.1}, 2D-CNNs (kernel size $3 \times 3$) are shared across future time steps, and separated 2D-CNNs (kernel size $1 \times 1$) are used to process the information from the residual path. We select the dimensions of pyramid decoder as $[192,96,48,2]$. We also split decoding heads for occupancy and flow to enable varying projections from the shared features and direct information paths. The output of the occupancy head for timestep $k$ is a two-dimensional vector for each grid cell denoting observed $\hat{O}_{k}^{b}$ and occluded $\hat{O}_{k}^{c}$ occupancy, while the output of the flow head is also a two-dimensional vector denoting the flow along $d_x$ and $d_y$ directions for $\hat{F}_{k}$.

\subsection{Loss Functions}
For better performance of joint occupancy and flow predictions, we modify the loss function in \cite{mahjourian2022occupancy}, but keep the binary probabilistic losses for occluded occupancy $\mathcal{L}_{occ}$ and observed occupancy $\mathcal{L}_{obs}$. Because the ground-truth samples are heavily imbalanced around zero (unoccupied area), we replace the cross-entropy loss with the focal loss $\mathcal{L}_{F}$\cite{lin2017focal}. For the flow-warped loss $\mathcal{L}_{\mathcal{W}}$, we use the ground-truth occupancy $O_{k-1}$ instead of the predicted one to stabilize flow training. The final multi-task learning objective sums up the following loss terms averaged by the height, width, and length of timesteps of the output:
\begin{equation}
\mathcal{L}=\frac{1}{hw{T}_{f}}\left(1000\mathcal{L}_{obs} + 1000\mathcal{L}_{occ} + 1000\mathcal{L}_{\mathcal{W}} + \mathcal{L}_{F}\right).
\end{equation}

\begin{table*}[htp]
\caption{Summary of the testing performance on the Waymo Occupancy and Flow Prediction Benchmark}
\centering
\setlength{\tabcolsep}{3.5mm}{
\begin{tabular}{l|cc|cc|c|cc}

\toprule
Evalutation Metrics       & \multicolumn{2}{c|}{Observed Occupancy} & \multicolumn{2}{c|}{Occluded Occupancy} & Flow & \multicolumn{2}{c}{Combined} \\\midrule
\textbf{Model}         & \textbf{AUC $\uparrow$}     & \textbf{Soft-IOU $\uparrow$}   & \textbf{AUC $\uparrow$}      & \textbf{Soft-IOU $\uparrow$}    & \textbf{EPE $\downarrow$} & \textbf{FT-AUC $\uparrow$} & \textbf{FT-Soft-IOU $\uparrow$} \\ \midrule
HorizonOccFlow(HOPE) \cite{hu2022hope}       & \textbf{0.803}   & 0.235               & 0.165             & 0.017                & 3.672        & \textbf{0.839}  & \textbf{0.633}       \\
Look Around \cite{lookaround} & 0.801   & 0.234               & 0.139             & 0.029                & \textbf{2.619}        & 0.825  & 0.549       \\
Temp-Q        & 0.757            & 0.393               & 0.171             & 0.041                & 3.308        & 0.778           & 0.465                \\
VectorFlow \cite{huang2022vectorflow}    & 0.755            & 0.488               & 0.174             & 0.045                & 3.583        & 0.767           & 0.531                \\
3D-STCNN \cite{he2019stcnn}     & 0.691            & 0.412               & 0.115             & 0.021                & 4.181        & 0.733           & 0.468                \\
Motionnet \cite{wang2017spatiotemporal}    & 0.694            & 0.411               & 0.141             & 0.032                & 4.275        & 0.732           & 0.469                \\
FTLS          & 0.618            & 0.318               & 0.085             & 0.019                & 9.612        & 0.689           & 0.431                \\
\midrule\rowcolor{mgray}
\textbf{Ours} & 0.778            & \textbf{0.491}      & \textbf{0.178}    &\textbf{ 0.045}                & 3.204        & 0.785           & 0.531               
\\\bottomrule
\end{tabular}
}
\label{table1}
\vspace{-0.4cm}
\end{table*}

\section{Experiments}
\subsection{Experimental Setup}
We employ the Waymo Open Motion dataset (WOMD) \cite{ettinger2021large} in the experiments, which consists of over 500,000 samples covering diverse real-world driving scenarios and dynamical interactions among traffic agents including vehicles, cyclists, and pedestrians. The historical agent states are sampled at 10Hz for the past one second ($T_h=10$), and the objective is to predict the occupancy and flow over the future 8 seconds at 1Hz ($T_f=8$). The rasterized image resolution for input and the output is $H,W=256$, representing an area of $80\times80 \ m^2$ in the real world; Hidden dimension is kept as $C=96$; The vector inputs $\mathcal{S}$ are sorted according to their current distances to the ego vehicle, and we keep a maximum of $n=64$ agents. The WOMD splits 485,568 samples for training and some scenarios of interest for validation and testing (4,400 each). 

To fairly evaluate the performance of our method, we follow the standard metrics proposed in the challenge \cite{mahjourian2022occupancy}. 1) Occupancy metrics: for $\hat{O}_{k}^{b}$ and $\hat{O}_{k}^{c}$, we measure \textbf{AUC} for the pair of precision-recall area values, and \textbf{Soft-IOU} for the overlapped area with ground-truth. 2) Flow metrics: \textbf{EPE} measures the mean L2 pixel distances of flow end-point error by $\hat{F}_{k}$. 3) Combined metrics: we measures the AUC and Soft-IOU for flow-traced occupancy according to Eq. \ref{e1} (\textbf{FT-AUC}, \textbf{FT-Soft-IOU}).

\subsection{Implementation details}
We choose GELU as the activation function in all encoders and ELU in the pyramid decoder. To mitigate over-fitting, dropout is added after each MLP layer and also in the image encoder, all with a dropout rate of 0.1. Due to the numerous size of data inputs and predictions, we use a distributed training strategy on 4 Tesla V100 GPUs with a total batch size of 16. Adam optimizer is used with an initial learning rate of 1e-4, and the learning rate decays by a factor of 50\% every 3 epochs. The total training epochs are set to 10.

\subsection{Quantitative Results}
\textbf{1) Performance on the benchmark:} Table \ref{table1} reports the testing performances of our proposed method against other state-of-the-art methods on the Waymo prediction benchmark. As of Aug 2022, our method has achieved \textbf{three best metrics}, i.e., Soft-IOU for both observed and occluded occupancy predictions, as well as the AUC for occluded occupancy. Moreover, our method achieves comparable performance to large pre-trained models, i.e., \textbf{HOPE} (\emph{Honorable Mention}) \cite{hu2022hope} and \textbf{Look Around} (\emph{$1^{st}$ place}) \cite{lookaround}, which both use very large image encoders pre-trained on ImageNet \cite{liu2021swin,sun2019deep}. Compared with the similar framework \textbf{VectorFlow} (\emph{$3^{rd}$ place}) \cite{huang2022vectorflow}, which also uses flattened visual features to attend to the vector features, our method performs better across a variety of metrics, i.e., over 2\% improvement in terms of occupancy metrics and 8\% for the flow prediction errors, as well as better combined metrics. Overall, the superior testing results of our method indicate that: 1) our method shows excellent capabilities in detecting the occurrence of traffic agents (Soft-IOU for detection); 2) the framework is more adaptive and capable of predicting the occluded (speculative) agents (high AUC), which cloud enhance the safety of downstream planning; 3) the proposed FG-MSA module described in Section \ref{FG} could improve the learning pipeline for flow and occupancy predictions.

\begin{figure}[htp]
    \centering
    \includegraphics[width=\linewidth]{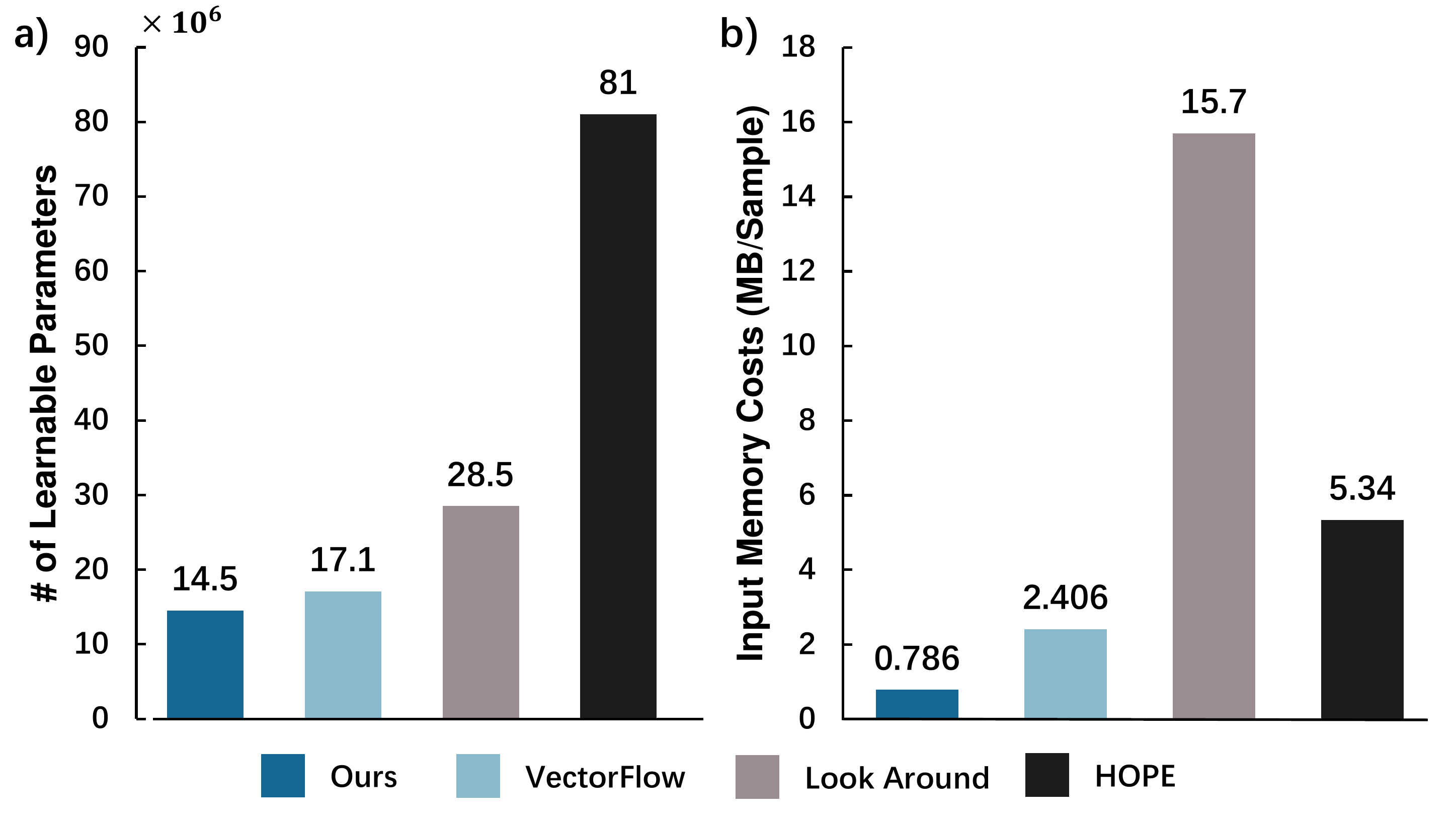}
    \caption{Comparison of \textbf{a)} learnable parameters for our method (full) and other baselines (encoders only); \textbf{b)} memory usage of input tensors for different models. Our method has much fewer network parameters and reduced memory usage.}    
    \label{fig5}
\end{figure}

\textbf{2) Computation efficiency:} To evaluate the computation efficiency of existing methods, we report the number of network parameters (Fig. \ref{fig5}(a)) and roughly calculate the memory usage of input tensors per sample (Fig. \ref{fig5}(b)) for each method. The calculation of tensor memory usage is explained as follows. We count 1 bit for occupancy at each grid cell (boolean type), 2 bytes for int type data, and 4 bytes for vectorized inputs (float type) per unit. The results in Fig. \ref{fig5}(b) show the outstanding memory efficiency of the proposed method, only 32\% of VectorFlow \cite{huang2022vectorflow} and 5\% of Look Around \cite{lookaround} in terms of memory usage. Likewise, in Fig. \ref{fig5}(a), comparing the number of network parameters of ours (full model) with other SOTA methods (encoders only), our model is much more concise, i.e., 18\% of HOPE \cite{hu2022hope} and 51\% of Look Around \cite{lookaround}, but is able to deliver competitive performance. Therefore, we can conclude that: 1) using unified large tensors combining all features consumes much more memory, and the proposed model can tackle this issue by combining visual images of occupancy and vectorized trajectories; 2) it is not necessary to use deep layers and encoding stages in visual encoders for the occupancy flow field prediction task, as our proposed method can deliver competitive results with only a few encoding stages.

\begin{figure*}[ht]
    \centering
    \includegraphics[width=\linewidth]{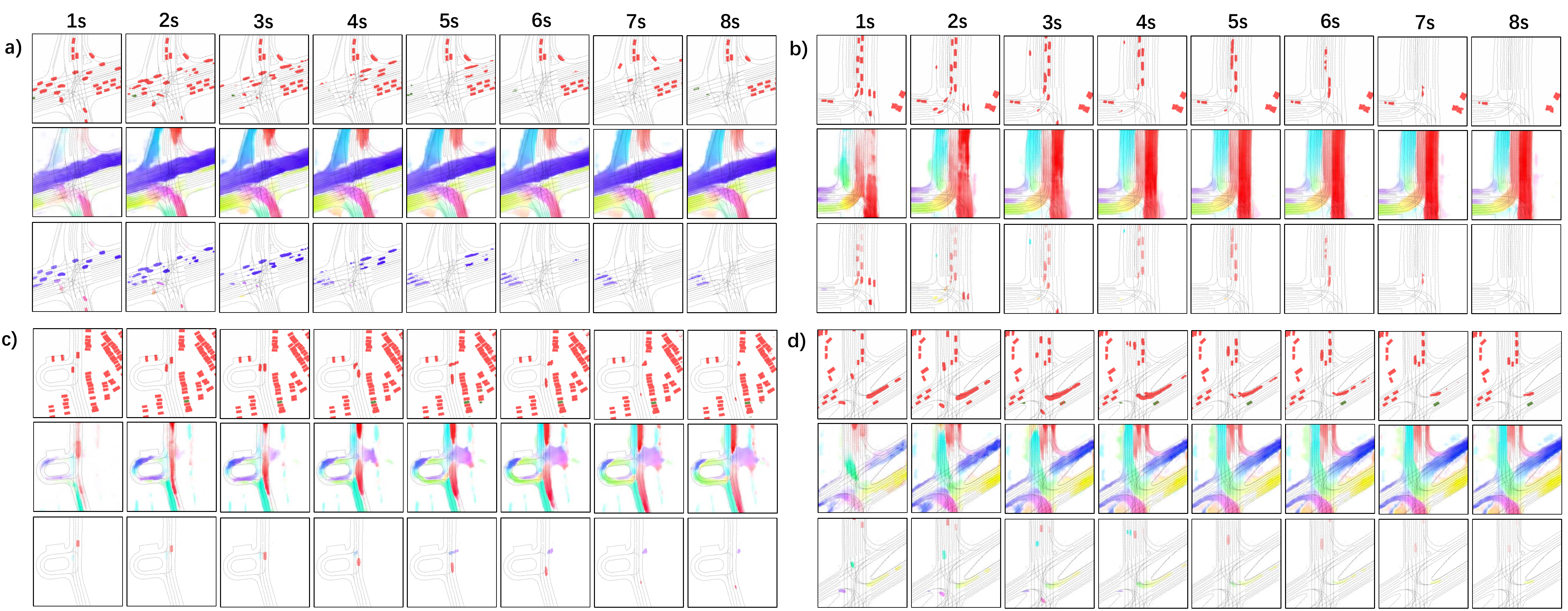}
    \caption{Qualitative results from the WOMD testing set. Each subplot displays the testing result of 1) observed (red) and occluded (green) occupancy, 2) backward flow, and 3) flow-traced occupancy. Some representative scenarios are listed: \textbf{a)} crossroad; \textbf{b)} T-intersection; \textbf{c)} parking lot with ring road; \textbf{d)} 5-way intersection.} 
    \label{fig:fig.6}
    \vspace{-0.5cm}
\end{figure*}

\subsection{Qualitative Results}
To intuitively evaluate the performance of our method, we visualize the testing results from several representative driving scenarios in Fig. \ref{fig:fig.6}. The results demonstrate that our method can perform effective and accurate occupancy forecasting for both dynamic (a, b, d) and static (c) traffic agents. Occlusion awareness is also manifested (a, c, d) even for popped-up agents (c, d). The flow-traced occupancy predictions further ensure the tractability of dynamic agents.

\begin{figure}[htp]
    \centering
    \includegraphics[width=\linewidth]{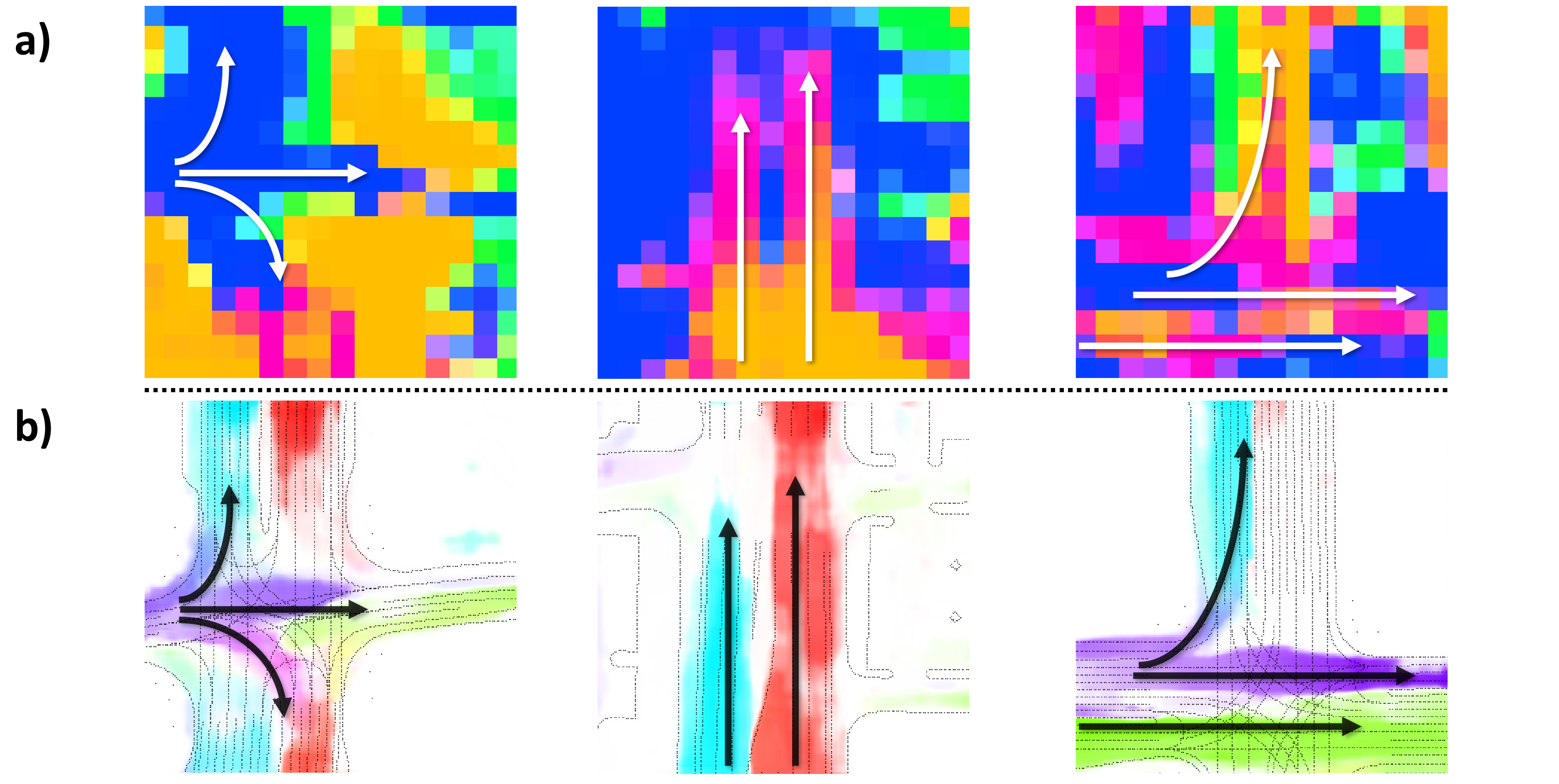}
    \caption{Comparison of \textbf{a)} flow offsets from FG-MSA and \textbf{b)} corresponding backward flow outputs. The similarity of flow trends in a) (white arrows) and b) (black arrows) indicates the effectiveness of the FG-MSA module's function to aggregate occupancy and flow features.}    
    \label{fig7}
\end{figure}

To investigate the performance improvement brought by the FG-MSA module, we compare the shape of the flow offsets inside the module (a) with flow outputs (b) in Fig. \ref{fig7}. The similar shapes of flow offsets and flow outputs suggest the function of the proposed FG-MSA module to guide the feature aggregation for both flow and occupancy. 

\subsection{Ablation Study}
We conduct an ablation study to investigate the influences of key modules in our proposed framework, i.e., the FG-MSA module and vectorized trajectories encoding and fusion module. Therefore, we train two ablated versions without the FG-MSA module and the vector-encoding branch respectively, and all the ablated models are validated on the same testing set. We report the AUC of occupancy metrics and the flow EPE in Table \ref{table2}, which shows that there is an overall performance drop in the two ablated models. Using purely visual-based encoding models significantly worsens the prediction performance due to the lack of agent motion information. Incorporating sparse agent motion information in the visual image, like in other baseline methods, can help mitigate this issue but consume a large number of memory usage and computing resources. On the other hand, we propose to use vectorized motion information and temporal cross-attention to fuse the visual and vector features can bring comparable performance but much less memory and computation usage. Compared with the ablated model without the FG-MSA module (\emph{$2^{nd}$ place} solution in the challenge), the current version shows significant improvements in observed AUC ($\uparrow3.5\%$) and flow EPE ($\downarrow9.9\%$). Because the FG-MSA module can better separate and aggregate the features from occupancy and flow and explicitly represents their relations (Eq. \ref{e1}), incorporating the FG-MSA module can further lift the model performance, especially for flow prediction.

\begin{table}[htbp]
\centering
\caption{Ablation study on FG-MSA and vector encoding}
\setlength{\tabcolsep}{2.5mm}{
\begin{tabular}{cc|cccc}
\toprule
Vector & FG-            & \textbf{Observed}     & \textbf{Occluded}     & \textbf{FT-}          & \textbf{Flow}         \\
Encoding & MSA & \textbf{AUC $\uparrow$} & \textbf{AUC $\uparrow$} & \textbf{AUC $\uparrow$} & \textbf{EPE $\downarrow$}  \\
\midrule
\XSolid    & \XSolid         & 0.741        & 0.138        & 0.751        & 3.712        \\
\Checkmark    & \XSolid         & 0.751        & 0.161        & 0.777        & 3.586        \\ \midrule
 \Checkmark   &    \Checkmark   & \textbf{0.778}        & \textbf{0.178}        & \textbf{0.785}        & \textbf{3.204}       \\ \bottomrule
\end{tabular}
\vspace{-0.1cm}
}

\label{table2}
\vspace{-0.4cm}
\end{table}

\section{Conclusions}
In this paper, we propose a multi-modal Hierarchical Transformer framework to forecast the occupancy flow field for autonomous driving. Multi-modal scene representation inputs, including visual features and vectorized motion trajectories, are separately encoded through carefully designed hierarchical Transformer modules, and both modalities are fused with temporal cross-attention. Moreover, the designed flow-guided attention module can better aggregate the flow and occupancy features via self-attention with explicit modeling of their mathematical relations. Comprehensive experiments conducted on the Waymo open dataset reveal the superior performance of the proposed method compared with SOTA models, even with a much smaller network. The ablation study indicates that adding the vectorized motion information fusion and flow-guided attention aggregation can significantly improve the prediction performance.

\bibliographystyle{IEEEtran}
\bibliography{b1}

\begin{thebibliography}{10}
\providecommand{\url}[1]{#1}
\csname url@samestyle\endcsname
\providecommand{\newblock}{\relax}
\providecommand{\bibinfo}[2]{#2}
\providecommand{\BIBentrySTDinterwordspacing}{\spaceskip=0pt\relax}
\providecommand{\BIBentryALTinterwordstretchfactor}{4}
\providecommand{\BIBentryALTinterwordspacing}{\spaceskip=\fontdimen2\font plus
\BIBentryALTinterwordstretchfactor\fontdimen3\font minus
  \fontdimen4\font\relax}
\providecommand{\BIBforeignlanguage}[2]{{%
\expandafter\ifx\csname l@#1\endcsname\relax
\typeout{** WARNING: IEEEtran.bst: No hyphenation pattern has been}%
\typeout{** loaded for the language `#1'. Using the pattern for}%
\typeout{** the default language instead.}%
\else
\language=\csname l@#1\endcsname
\fi
#2}}
\providecommand{\BIBdecl}{\relax}
\BIBdecl

\bibitem{mozaffari2020deep}
S.~Mozaffari, O.~Y. Al-Jarrah, M.~Dianati, P.~Jennings, and A.~Mouzakitis,
  ``Deep learning-based vehicle behavior prediction for autonomous driving
  applications: A review,'' \emph{IEEE Transactions on Intelligent
  Transportation Systems}, vol.~23, no.~1, pp. 33--47, 2020.

\bibitem{huang2021driving}
Z.~Huang, J.~Wu, and C.~Lv, ``Driving behavior modeling using naturalistic
  human driving data with inverse reinforcement learning,'' \emph{IEEE
  Transactions on Intelligent Transportation Systems}, 2021.

\bibitem{huang2022differentiable}
Z.~Huang, H.~Liu, J.~Wu, and C.~Lv, ``Differentiable integrated motion
  prediction and planning with learnable cost function for autonomous
  driving,'' \emph{arXiv preprint arXiv:2207.10422}, 2022.

\bibitem{mo2022multi}
X.~Mo, Z.~Huang, Y.~Xing, and C.~Lv, ``Multi-agent trajectory prediction with
  heterogeneous edge-enhanced graph attention network,'' \emph{IEEE
  Transactions on Intelligent Transportation Systems}, 2022.

\bibitem{mo2020interaction}
X.~Mo, Y.~Xing, and C.~Lv, ``Interaction-aware trajectory prediction of
  connected vehicles using cnn-lstm networks,'' in \emph{IECON 2020 The 46th
  Annual Conference of the IEEE Industrial Electronics Society}.\hskip 1em plus
  0.5em minus 0.4em\relax IEEE, 2020, pp. 5057--5062.

\bibitem{djuric2020uncertainty}
N.~Djuric, V.~Radosavljevic, H.~Cui, T.~Nguyen, F.-C. Chou, T.-H. Lin,
  N.~Singh, and J.~Schneider, ``Uncertainty-aware short-term motion prediction
  of traffic actors for autonomous driving,'' in \emph{Proceedings of the
  IEEE/CVF Winter Conference on Applications of Computer Vision}, 2020, pp.
  2095--2104.

\bibitem{mahjourian2022occupancy}
R.~Mahjourian, J.~Kim, Y.~Chai, M.~Tan, B.~Sapp, and D.~Anguelov, ``Occupancy
  flow fields for motion forecasting in autonomous driving,'' \emph{IEEE
  Robotics and Automation Letters}, vol.~7, no.~2, pp. 5639--5646, 2022.

\bibitem{hu2022hope}
Y.~Hu, W.~Shao, B.~Jiang, J.~Chen, S.~Chai, Z.~Yang, J.~Qian, H.~Zhou, and
  Q.~Liu, ``Hope: Hierarchical spatial-temporal network for occupancy flow
  prediction,'' \emph{arXiv preprint arXiv:2206.10118}, 2022.

\bibitem{hoermann2018dynamic}
S.~Hoermann, M.~Bach, and K.~Dietmayer, ``Dynamic occupancy grid prediction for
  urban autonomous driving: A deep learning approach with fully automatic
  labeling,'' in \emph{2018 IEEE International Conference on Robotics and
  Automation (ICRA)}.\hskip 1em plus 0.5em minus 0.4em\relax IEEE, 2018, pp.
  2056--2063.

\bibitem{ngiam2021scene}
J.~Ngiam, B.~Caine, V.~Vasudevan, Z.~Zhang, H.-T.~L. Chiang, J.~Ling,
  R.~Roelofs, A.~Bewley, C.~Liu, A.~Venugopal \emph{et~al.}, ``Scene
  transformer: A unified architecture for predicting multiple agent
  trajectories,'' \emph{arXiv preprint arXiv:2106.08417}, 2021.

\bibitem{varadarajan2022multipath++}
B.~Varadarajan, A.~Hefny, A.~Srivastava, K.~S. Refaat, N.~Nayakanti,
  A.~Cornman, K.~Chen, B.~Douillard, C.~P. Lam, D.~Anguelov \emph{et~al.},
  ``Multipath++: Efficient information fusion and trajectory aggregation for
  behavior prediction,'' in \emph{2022 International Conference on Robotics and
  Automation (ICRA)}.\hskip 1em plus 0.5em minus 0.4em\relax IEEE, 2022, pp.
  7814--7821.

\bibitem{gu2021densetnt}
J.~Gu, C.~Sun, and H.~Zhao, ``Densetnt: End-to-end trajectory prediction from
  dense goal sets,'' in \emph{Proceedings of the IEEE/CVF International
  Conference on Computer Vision}, 2021, pp. 15\,303--15\,312.

\bibitem{kim2022stopnet}
J.~Kim, R.~Mahjourian, S.~Ettinger, M.~Bansal, B.~White, B.~Sapp, and
  D.~Anguelov, ``Stopnet: Scalable trajectory and occupancy prediction for
  urban autonomous driving,'' \emph{arXiv preprint arXiv:2206.00991}, 2022.

\bibitem{lang2019pointpillars}
A.~H. Lang, S.~Vora, H.~Caesar, L.~Zhou, J.~Yang, and O.~Beijbom,
  ``Pointpillars: Fast encoders for object detection from point clouds,'' in
  \emph{Proceedings of the IEEE/CVF conference on computer vision and pattern
  recognition}, 2019, pp. 12\,697--12\,705.

\bibitem{liu2021swin}
Z.~Liu, Y.~Lin, Y.~Cao, H.~Hu, Y.~Wei, Z.~Zhang, S.~Lin, and B.~Guo, ``Swin
  transformer: Hierarchical vision transformer using shifted windows,'' in
  \emph{Proceedings of the IEEE/CVF International Conference on Computer
  Vision}, 2021, pp. 10\,012--10\,022.

\bibitem{gao2020vectornet}
J.~Gao, C.~Sun, H.~Zhao, Y.~Shen, D.~Anguelov, C.~Li, and C.~Schmid,
  ``Vectornet: Encoding hd maps and agent dynamics from vectorized
  representation,'' in \emph{Proceedings of the IEEE/CVF Conference on Computer
  Vision and Pattern Recognition}, 2020, pp. 11\,525--11\,533.

\bibitem{huang2022multi}
Z.~Huang, X.~Mo, and C.~Lv, ``Multi-modal motion prediction with
  transformer-based neural network for autonomous driving,'' in \emph{2022
  International Conference on Robotics and Automation (ICRA)}.\hskip 1em plus
  0.5em minus 0.4em\relax IEEE, 2022, pp. 2605--2611.

\bibitem{dosovitskiy2020image}
A.~Dosovitskiy, L.~Beyer, A.~Kolesnikov, D.~Weissenborn, X.~Zhai,
  T.~Unterthiner, M.~Dehghani, M.~Minderer, G.~Heigold, S.~Gelly \emph{et~al.},
  ``An image is worth 16x16 words: Transformers for image recognition at
  scale,'' \emph{arXiv preprint arXiv:2010.11929}, 2020.

\bibitem{xia2022vision}
Z.~Xia, X.~Pan, S.~Song, L.~E. Li, and G.~Huang, ``Vision transformer with
  deformable attention,'' in \emph{Proceedings of the IEEE/CVF Conference on
  Computer Vision and Pattern Recognition}, 2022, pp. 4794--4803.

\bibitem{bansal2018chauffeurnet}
M.~Bansal, A.~Krizhevsky, and A.~Ogale, ``Chauffeurnet: Learning to drive by
  imitating the best and synthesizing the worst,'' \emph{arXiv preprint
  arXiv:1812.03079}, 2018.

\bibitem{hu2021fiery}
A.~Hu, Z.~Murez, N.~Mohan, S.~Dudas, J.~Hawke, V.~Badrinarayanan, R.~Cipolla,
  and A.~Kendall, ``Fiery: Future instance prediction in bird's-eye view from
  surround monocular cameras,'' in \emph{Proceedings of the IEEE/CVF
  International Conference on Computer Vision}, 2021, pp. 15\,273--15\,282.

\bibitem{huang2022vectorflow}
X.~Huang, X.~Tian, J.~Gu, Q.~Sun, and H.~Zhao, ``Vectorflow: Combining images
  and vectors for traffic occupancy and flow prediction,'' \emph{arXiv preprint
  arXiv:2208.04530}, 2022.

\bibitem{huang2022recoat}
Z.~Huang, X.~Mo, and C.~Lv, ``Recoat: A deep learning-based framework for
  multi-modal motion prediction in autonomous driving application,''
  \emph{arXiv preprint arXiv:2207.00726}, 2022.

\bibitem{lin2017feature}
T.-Y. Lin, P.~Doll{\'a}r, R.~Girshick, K.~He, B.~Hariharan, and S.~Belongie,
  ``Feature pyramid networks for object detection,'' in \emph{Proceedings of
  the IEEE conference on computer vision and pattern recognition}, 2017, pp.
  2117--2125.

\bibitem{lin2017focal}
T.-Y. Lin, P.~Goyal, R.~Girshick, K.~He, and P.~Doll{\'a}r, ``Focal loss for
  dense object detection,'' in \emph{Proceedings of the IEEE international
  conference on computer vision}, 2017, pp. 2980--2988.

\bibitem{lookaround}
\BIBentryALTinterwordspacing
P.~Dmytro, ``Waymo open dataset occupancy and flow prediction challenge
  solution: Look around,'' 2022. [Online]. Available:
  \url{https://storage.googleapis.com/waymo-uploads/files/research/OccupancyFlow/Dmytro1.pdf}
\BIBentrySTDinterwordspacing

\bibitem{he2019stcnn}
Z.~He, C.-Y. Chow, and J.-D. Zhang, ``Stcnn: A spatio-temporal convolutional
  neural network for long-term traffic prediction,'' in \emph{2019 20th IEEE
  International Conference on Mobile Data Management (MDM)}.\hskip 1em plus
  0.5em minus 0.4em\relax IEEE, 2019, pp. 226--233.

\bibitem{wang2017spatiotemporal}
Y.~Wang, M.~Long, J.~Wang, and P.~S. Yu, ``Spatiotemporal pyramid network for
  video action recognition,'' in \emph{Proceedings of the IEEE conference on
  Computer Vision and Pattern Recognition}, 2017, pp. 1529--1538.

\bibitem{ettinger2021large}
S.~Ettinger, S.~Cheng, B.~Caine, C.~Liu, H.~Zhao, S.~Pradhan, Y.~Chai, B.~Sapp,
  C.~R. Qi, Y.~Zhou \emph{et~al.}, ``Large scale interactive motion forecasting
  for autonomous driving: The waymo open motion dataset,'' in \emph{Proceedings
  of the IEEE/CVF International Conference on Computer Vision}, 2021, pp.
  9710--9719.

\bibitem{sun2019deep}
K.~Sun, B.~Xiao, D.~Liu, and J.~Wang, ``Deep high-resolution representation
  learning for human pose estimation,'' in \emph{Proceedings of the IEEE/CVF
  conference on computer vision and pattern recognition}, 2019, pp. 5693--5703.

\end{thebibliography}
\end{document}